%% file: conference_041818.tex
\documentclass[conference]{IEEEtran}
\IEEEoverridecommandlockouts
\usepackage{cite}
\usepackage{amsmath,amssymb,amsfonts}
\usepackage{graphicx}
\usepackage{textcomp}
\usepackage{algorithm}
\usepackage{algpseudocode}
\usepackage{color,soul}

\usepackage{verbatim}
\usepackage[breaklinks=true]{hyperref}
\usepackage{cite}
\usepackage{textcomp}
\usepackage[table]{xcolor}
\usepackage{enumitem}
\usepackage{multicol}
\usepackage{multirow}
\usepackage{gensymb}

\usepackage{algorithm}
\usepackage{algpseudocode}

\usepackage{bm}

\usepackage{booktabs,tabularx,enumitem,ragged2e}
\usepackage{colortbl}
\usepackage[utf8]{inputenc}

\usepackage{xcolor}
\newcommand\sudip[1]{\textcolor{blue}{Sudip : #1}}

\usepackage{array}
\usepackage{multirow}
\usepackage{graphicx,array}
\usepackage[table]{xcolor}

\usepackage[singlelinecheck=false]{caption}

\usepackage{romannum}

\usepackage{caption}
\usepackage{subcaption}

\def\BibTeX{{\rm B\kern-.05em{\sc i\kern-.025em b}\kern-.08em
    T\kern-.1667em\lower.7ex\hbox{E}\kern-.125emX}}
\begin{document}

\title{A Temporal Anomaly Detection System for Vehicles utilizing Functional Working Groups and Sensor Channels}

%


\author{\IEEEauthorblockN{Subash Neupane*, Ivan A. Fernandez*\thanks{* These two authors contributed equally.}, Wilson Patterson, Sudip Mittal, Shahram Rahimi}

Department of Computer Science \& Engineering \\ Mississippi State University, Mississippi, USA, \\(email: \{sn922, iaf28, wep104\}@msstate.edu, \{mittal, rahimi\}@cse.msstate.edu)

}

\maketitle

\begin{abstract}

A modern vehicle fitted with sensors, actuators, and Electronic Control Units (ECUs) can be divided into several operational subsystems called Functional Working Groups (FWGs). Examples of these FWGs include the engine system, transmission, fuel system, brakes, etc. Each FWG has associated sensor-channels that gauge vehicular operating conditions. 
This data rich environment is conducive to the development of Predictive Maintenance (PdM) technologies. Undercutting various PdM technologies is the need for robust anomaly detection models that can identify
events or observations which deviate significantly from the
majority of the data and do not conform to a well defined
notion of normal vehicular operational behavior.
In this paper, we introduce the Vehicle Performance, Reliability, and Operations (VePRO) dataset and use it to create a multi-phased approach to anomaly detection. 
Utilizing Temporal Convolution Networks (TCN), our anomaly detection system can achieve 96\% detection accuracy and accurately predicts 91\% of true anomalies. The performance of our anomaly detection system improves when sensor channels from multiple FWGs are utilized.

\end{abstract}


\section{Introduction}


Vehicles can be considered a specialized form of Cyber Physical Systems (CPS) equipped with a variety of sensors that generate a volume of operational data. Recent advancements in hardware has made available unprecedented computational power via the use of modern processors and GPUs.  Artificial Intelligence / Machine learning technologies running on this hardware, can utilize these large vehicular datasets to build intelligent systems. These systems can help detect abnormal sensor behaviour and identify precursors of various sensor channel faults, monitor and predict the progression of faults, and provide decision-support to develop maintenance schedules. One concept that has risen to prominence both in academia and industry is \emph{Predictive Maintenance (PdM)} \cite{ran2019survey,zhang2019data,zonta2020predictive,carvalho2019systematic}. However, PdM is complex and expensive to implement \cite{ran2019survey}. Undercutting the domain of PdM, is the need for robust \textit{anomaly detection}. AI based anomaly detection systems can identify events or observations that deviate significantly from the majority of the data and do not conform to a well defined notion of normal behavior \cite{chalapathy2019deep, kwon2019survey, ma2021comprehensive, chandola2009anomaly}. 

In vehicular systems, several factors are attributed to the deterioration of vehicle components over time. For example, engine and transmission wear is a direct result of massive heat and vibration produced. This deterioration of components or sub-components can result in abnormal sensor channel behavior signaling impending failure. However, early detection of abnormal patterns and timely scheduling of maintenance events can reduce risk to the underlying processes while increasing lifespan, reliability, and availability, thereby avoiding unplanned downtime and minimizing repair costs \cite{ran2019survey}.


When it comes to a modern vehicle, systems have shifted from being purely mechanical to being comprised of Electronic Control Units (ECU), Sensor, and Actuators. These components work together to ensure normal operation of a vehicle. An ECU waits for a signal from the gas pedal sensor to communicate with a fuel pump, instructing it to deliver more fuel to the internal combustion engine. A modern vehicle is an amalgamation of these individual pieces of machinery working together to complete the desired functionality. We can consider a vehicle to be comprised of various operational subsystems called ``\textit{Functional Working Groups}" (FWGs) \cite{alizadeh2021vehicle}. These include the \emph{engine system, transmission, fuel system, brakes system,} etc. 
Each FWG is comprised of multiple sensor channels that allow a mechanic to infer the condition of an individual mechanical part and recommend mitigating service. We can say that a vehicle is operating under normal conditions if all individual FWG sensor channel values are gauging normally. 
Abnormal channel values may be indicative of a future impending failure. 

To combat this problem, we present in this paper a temporal anomaly detection system that focuses on operational data of 4 vehicular FWGs and associated sensor channels.
Our anomaly detection system is a multi-phased approach utilizing Temporal Convolution Networks (TCN) that learns the underlying relationships between FWG sensor channels. The intuition behind using TCN as a predictor model is that it is suitable for modeling sequential data and the use of flexible receptive fields \cite{he2019temporal}. In addition, TCN based models are computationally efficient as compared to Long short-term memory (LSTM) methods and possess representational capability to achieve robust and superior prediction performance \cite{bai2018empirical}. We leverage the Vehicle Performance, Reliability, and Operations (VePRO) dataset created by the United States (U.S.) Army Corps of Engineers and Mississippi State University (MSU). Our anomaly detection model performs time series forecasting by taking as an input, {multi-sensor}  values for the current time window and outputs a prediction for the subsequent observation. 
{The final decision about an anomaly is made by comparing the prediction to the actual observation.}



Major contributions of this paper include:
\begin{itemize}
    \item We provide a comprehensive description of the Vehicle Performance, Reliability, and Operations (VePRO) dataset. The dataset includes the normal driving data for a fleet of vehicles. We explain and utilize 4 vehicular Functional Working Groups such as engine, transmission, fuel, and brakes, along with the associated sensor channels.  
    \item We develop a TCN model using a sequence-to-sequence learning approach. The model performs time series forecasting by operating on a multi-channel sequence for the current time window and outputs a prediction for the subsequent observation.
    \item We identify anomalies in FWG sensor channels by comparing the prediction made by the TCN model with the actual observation. We then check robustness of our approach by creating three anomalous test scenarios to investigate the behavior of different sensor channels that affect {\emph{FuelRate}}.
\end{itemize}

The rest of the paper is structured as follows: Section \ref{background}, discusses the model used in the study and provides a brief overview of the current state-of-the-art. Dataset used in the study has been described in Section \ref{dataset_des}. Section \ref{architecture}, gives an overview of our multi-phased architecture. In Section \ref{evaluation}, we present the evaluation criteria and our results. Finally, Section \ref{conclusion} summarizes and concludes the paper.

\section{Background \& Related Work}
\label{background}
In this section, we present some background information and some related works on Temporal Convolution Network and Anomaly Detection.

\subsection{Temporal Convolution Network}\label{tcnback}

\begin{figure}[h]
    \centering
    \includegraphics[scale=.7]{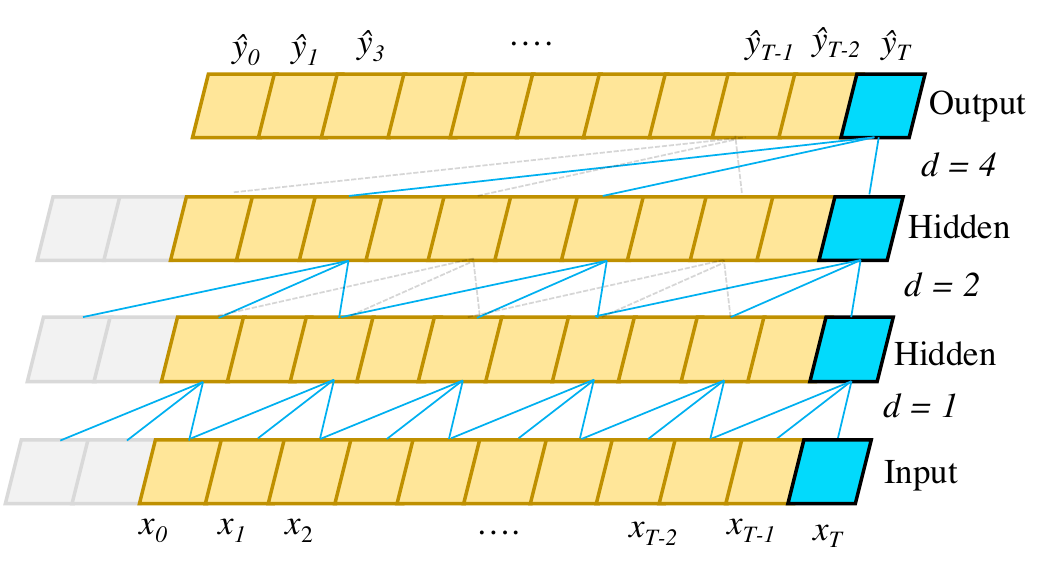}
    \caption{A causal dilated network with dilated factors $d$ = 1,2, 4 and a kernel size of 3. The first hidden layer has a dilation factor of 2{,} and the second hidden layer has a dilation factor of 4. The receptive field can accommodate all input sequence values \cite{bai2018empirical}.}
        \label{fig:convolution}
\end{figure} 
\begin{figure}[h]
    \centering
    \includegraphics[scale=.80]{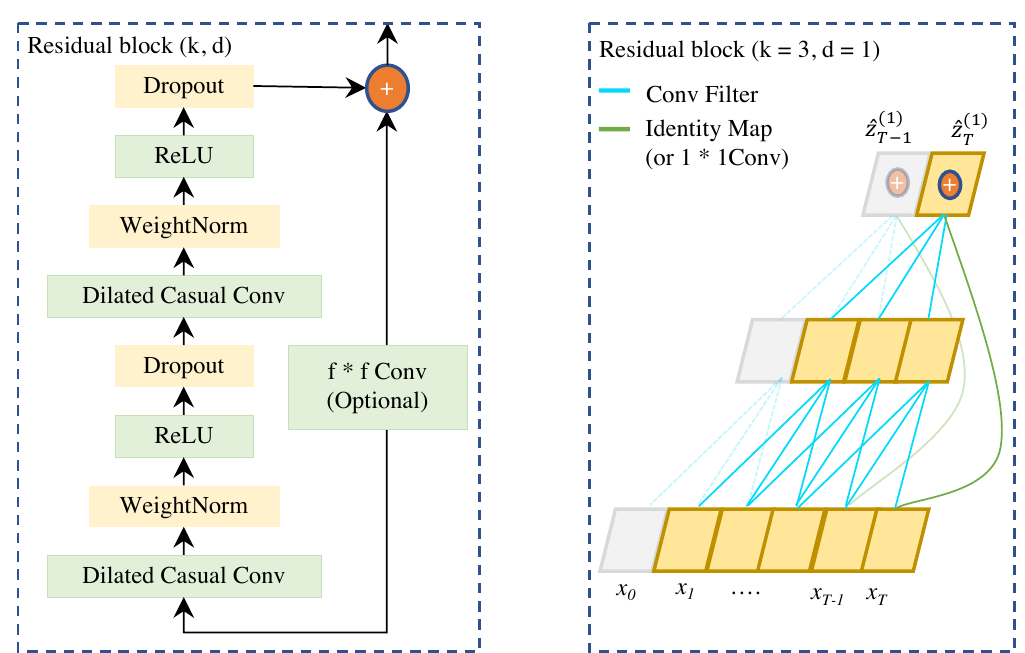}
    \caption{{The figure on the left shows a TCN residual block with a dilation factor of $d$ and a filter of $k$.} When the dimensions of the residual input and output differ, a $1 \times 1$ convolution is added. The figure on the right is an example of residual connections in a TCN with $d$ = 1 and $k$ = 3. In this figure, the green lines show identity mappings, while the blue lines show filters in the residual function \cite{bai2018empirical}.}
    \label{fig:residual_connection}
    \vspace{-4mm}
\end{figure} 

TCN  \cite{lea2017temporal}, is a time series architecture that utilizes one-dimensional convolutional layers. 
{It is trained to forecast the next sequence of time series input. Consider a scenario in which we are given a set of time series inputs for a window size $t$, i.e. $x{_0}$, $x{_1}$, $x{_2}$,..., $x{_t}$ and we wish to predict some corresponding output $y{_{t+1}}$, $y{_{t+2}}$, $y{_{t+3}}$,..., $y{_{t+t}}$ for the next time window of size $t$. The key limitation is that when predicting the outputs $y$ values for the next window size $t$, it can only use the previously observed inputs: $x_0$, $x_1$, $x_2$,..., $x_t$.} TCNs adhere to the following two fundamental principles \cite{bai2018empirical}: 

\begin{enumerate}
    \item The architecture employs a 1D fully connected convolutional network and transforms any dimensional length of input sequence into a sequence with the same dimensional length \cite{long2015fully}.
    \item The convolutions are causal, which means that there is no information leakage from the future to the past \cite{alla2019beginning}. 
\end{enumerate}

To allow for a long effective history size and a deep network, two important concepts are introduced as part of the TCN architectural elements, which we describe in the following sub sections.
\subsubsection{Dilated Convolution}
When working with time series data, it is often assumed that the model network can memorize long-term information. A simple causal convolution can only examine a history with a size proportional to the network's depth. This makes it difficult to apply causal convolution to sequence tasks, particularly those requiring a longer historical context. To address this issue, dilated convolution, originally proposed by Oored et al. \cite{oord2016wavenet} that enables exponentially large receptive fields is employed \cite{yu2015multi}. 

For a sequence of input $x_0$, $x_1$, $x_2$,..., $x_t$, and k-sized convolution kernel $f$, the dilated operation $F$ in element $s$ of the sequence is defined as 
\begin{equation}
F(s)=\left(\mathbf{x} *_d f\right)(s)=\sum_{i=0}^{k-1} f(i) \cdot \mathbf{x}_{s-d \cdot i}
\end{equation}
\noindent Where $d$ is dilation factor, $k$ is filter size, and $s-d \cdot i$ accounts for the direction of the past. Figure \ref{fig:convolution} illustrates the dilated-causal convolutions with dilation factors $d$ = 1, 2, and 4.

\subsubsection{Residual Connections}
Figure \ref{fig:convolution}, shows how increasing network depth or extending network width can increase network performance. However, as network depth increases, it becomes more difficult to train the network, and network performance rapidly degrades. To address this issue, residual connections are applied to TCN \cite{he2016deep}. This utilizes skip connections throughout the network to accelerate the training process and prevent the vanishing gradient problem. A residual block for baseline TCN is depicted in Figure \ref{fig:residual_connection}.

\subsection{Anomaly Detection}

\begin{figure*}[htbp]
    \centering
    \includegraphics[scale=0.8]{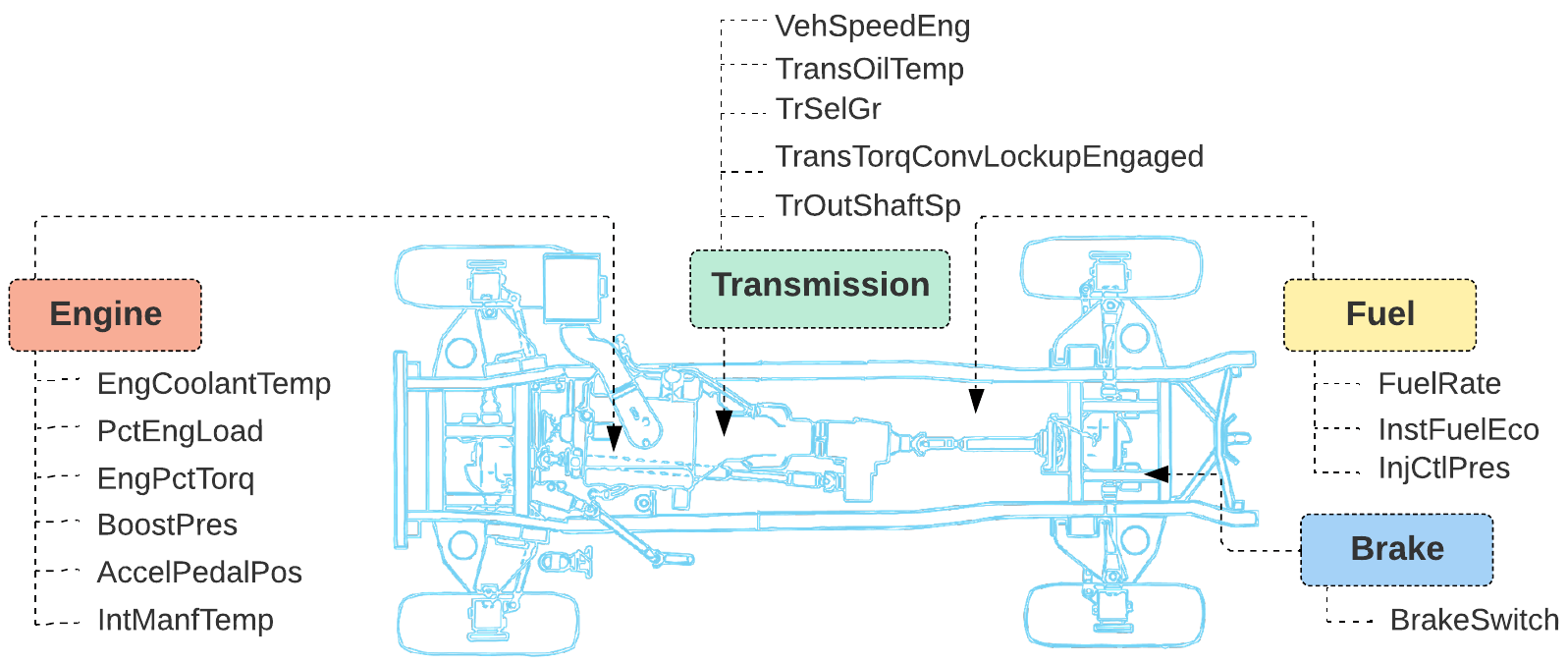}
    \caption{A graphical representation of the various FWG in a vehicle, including the \emph{engine, transmission, fuel}, and \emph{brake}, as well as the sensor channels for each of these FWG \cite{CoverHMMWVTM}.}
        \label{fig:vechicle_subsystems_diagram}
        
        \vspace{-4mm}
\end{figure*}

\input{Table/channel_summary}

Liu et al.\cite{liu2019anomaly} propose a TCN-Gaussian mixture model (GMM) model for an online anomaly detection time series dataset. First, the time series features are extracted by a dilated TCN. Then their distribution is estimated using a GMM. TCN-GMM measures the similarity between features and clusters using the Mahalanobis distance. The threshold \emph{T} is compared to the model's measure of similarity to see if the current state is abnormal.

On the other hand, the authors of \cite{he2019temporal} use a multi-stage TCN (to improve performance) and a multivariate Gaussian distribution to estimate the distribution of prediction errors rather than the features of training data. The anomaly score is then calculated by computing the Mahalanobis distance (MD) between the current prediction error and the pre-estimated error distribution. Three real-world datasets, like electrocardiograms, a dataset of 2-D gestures, and a dataset from the space shuttle, are used to validate the approach.

In another work, using Temporal Convolutional Neural Attention (TCNA) networks, the authors of \cite{thiruloga2022tenet} present a novel anomaly detection framework called TENET for automotive cyber-physical systems. The research introduces a metric called the ``divergence score'' that evaluates the discrepancy between the expected signal value and the actual signal value. 
{Their method employs a robust quantitative metric and classifier, as well as learned relationships, to detect abnormal patterns.} 

A comparison of the effectiveness of TCN and LSTM methods in detecting anomalies in time series data has been proposed \cite{gopali2021comparison}. 
Experimental results demonstrated that the TCN method outperformed its LSTM equivalent in terms of recall 
and F1 score. 
In addition, it is noted that the TCN model required fewer training iterations. A similar work that evaluates the performance of the TCN approach by comparing its results with the LSTM approach in detecting anomalies of spacecraft telemetry time series data transmitted from the on-orbit satellite is proposed in \cite{wang2021anomaly}. 
 
Mo et al. \cite{mo2021unsupervised} present a TCN-based unsupervised anomaly identification approach for time series data that includes seasonality and trend. In contrast to the typical method of setting a global threshold for {anomaly} detection, the authors developed a novel approach capable of detecting seasonal and trend-driven local {anomalies}.

To build an anomaly detection system for vehicles, Alizadeh et al. \cite{alizadeh2022hybrid} created a hybrid AutoRegressive Integrated Moving Average-Wavelet Neural Networks (ARIMA-WNN) model for predicting the behavior of the operating vehicle and detecting anomalous states based on multiple channel time series data. 

{In our past work, we have published a Hidden Markov Model (HMM) based survey on machinery fault prediction} \cite{9659838} and have also built a number of anomaly detection systems for vehicles, smart homes, cyber-physical systems, etc. These systems utilize both non-neural network methods and various deep learning approaches \cite{narayanan2016obd_securealert, chukkapalli2021privacy, ramapatruni2019anomaly, narayanan2016using,nair2015using}. For example, in \cite{narayanan2016obd_securealert}, we presented an alert system based on HMM, a stochastic model that follows the Markov principle, to detect anomalous states of real data acquired from an operating vehicle. This research looked at the speed, load, engine coolant temperature, and other physical sensor values of vehicles made by different automotive manufacturers.

Kang et al. \cite{kang2021anomaly} also proposed a method to detect anomalies in the brake operating unit (BOU) of subway trains using a type of LSTM auto-encoder. 
The authors of \cite{cheifetz2011pattern} propose a pattern recognition technique for detecting anomalies in transit bus braking systems. This technique employs many sensor data points, such as vehicle speed, transmission gear ratios, acceleration pedal position, and the brake switch, to detect irregularities in brake systems. Bussey et al. \cite{bussey2014case} present a case study on the detection of anomalies involving low oil pressure in commercial vehicle data. {The authors studied a number of data-driven models that were based on heuristics to describe the relationship between three main engine channels: oil pressure, oil temperature, and engine speed.} 

\section{Dataset Description}
\label{dataset_des}
The Vehicle Performance, Reliability, and Operations (VePRO) program is a collaboration between the United States (U.S.) Army Corps of Engineers and Mississippi State University (MSU). The VePRO dataset contains operational data from a fleet of vehicles{,} including time series sensor measurements, detected faults, and maintenance reports. 
Each vehicle in the dataset possesses a minimum of 106 sensor channels.
The multi-channel sensor data was collected daily over a period of 21 months between 2012 and 2014. The dataset is scheduled to be released publicly after detailed anonymization.

A subset of data with 15 sensor channels and one additional time channel \emph{UTC\_1Hz} is selected for our anomaly detection problem as per the recommendation of a group of mechanical engineers working on the same dataset. The selected 15 sensor channels belonging to the four main FWGs of a vehicle are the basis for the data analysis, model training, and studies performed in this paper. \emph{UTC\_1Hz} sensor channel records the date and timestamp for every observation of data. It is collected at a frequency of 1 sample per second, starting at the timestamp value when the engine is switched on.

\begin{figure}[h]
    \centering
    \begin{subfigure}[b]{0.5\textwidth}
    \includegraphics[scale=0.7]{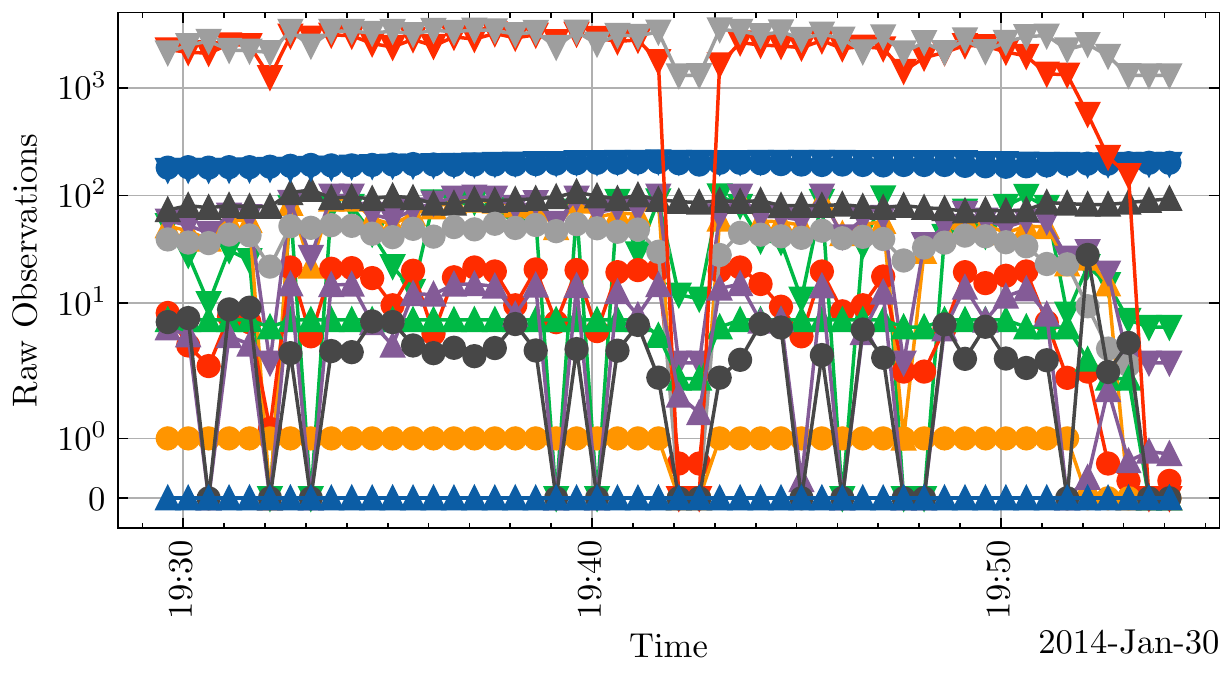}
    \end{subfigure}
    \begin{subfigure}[b]{0.3\textwidth}
    \includegraphics[scale=0.6]{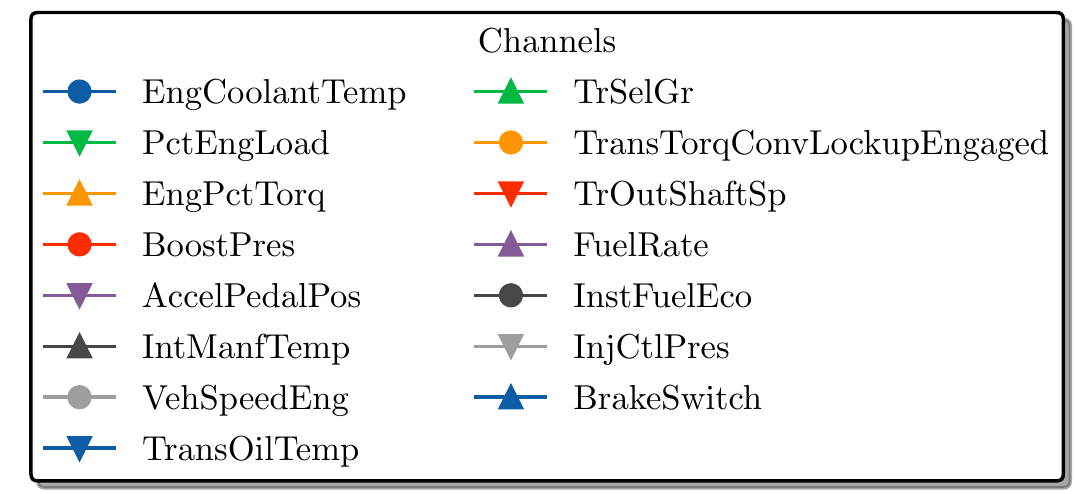}
    \end{subfigure}
    \caption{Snapshot of VePro Dataset raw observations for a period of 25 minutes on 2014-Jan-30.}
    \label{fig:sequence_example}
\end{figure}

The selected vehicular sensor channels can be visualized in Figure \ref{fig:vechicle_subsystems_diagram} and a summary of these channels is provided in Table \ref{table:features_description}. 
These channels specifically deal with certain ECUs installed in different vehicular FWGs, such as \emph{engine, transmission, fuel,} and \emph{brake}.  Figure \ref{fig:sequence_example} illustrates raw observations for a window of 25 minutes on 2014-Jan-30. Next, we describe each of the selected FWGs and associated sensor channels in detail. 
 
\subsection{Engine FWG and Sensor Channels}

The engine FWG converts energy from combustion to mechanical work. It includes the following sensor channels: 

\subsubsection{EngCoolantTemp (Engine Coolant Temperature)} 


The \emph{EngCoolantTemp} sensor channel measures the temperature of the engine's liquid coolant. 
The sensor operates by sensing the temperature emitted by the thermostat or the coolant itself. 
The vehicle's computer will use this information to keep the engine at the optimal temperature.

\subsubsection{PctEngLoad (Percent Engine Load)} The engine load determines the demand placed on a motor's ability to generate power. 
When the engine load increases, the engine speed decreases. To compensate for this decrease in engine speed, additional fuel is injected into the engine. 
\emph{PctEngLoad} sensor channel records fractional power of an engine compared to its maximum manufacturer's design capacity at various engine conditions, such as a vehicle in motion or a vehicle at idle.
\subsubsection{EngPctTorq (Engine Percent Torque)} Torque is commonly referred to as `twisting or turning force'. 
In the context of automobiles, it measures the rotational force applied by the piston to the crankshaft of the engine. The \emph{EngPctTorq} sensor channel is the calculated output torque of the engine. The data is recorded as a percentage of net engine torque \cite{rohrer2018evaluation}.

\subsubsection{BoostPres (Turbo Boost Pressure)} The \emph{BoostPres} sensor channel indicates the air pressure information at any given moment. It provides information regarding air pressure and air-to-fuel ratios for regulating engine performance. The ECU then uses boost and air density information in the vehicle’s intake manifold to determine how much fuel is needed in the car's combustion chamber so that the air-fuel mixture is at its best.

\subsubsection{AccelPedalPos (Accelerator Pedal Position)} The \emph{AccelPedalPos} sensor channel captures information about the acceleration, deceleration, and steady-state condition of a vehicle. It monitors the accelerator pedal's position and relays that information to the ECU. The computer opens the throttle body to increase the flow of fuel to the engine when the pedal is pushed forward. On the other hand, when the pedal is released, the throttle body is closed, and the flow of fuel is reduced.

\subsubsection{IntManfTemp (Intake Manifold Temperature)} The \emph{IntManfTemp} sensor channel indicates the temperature of the air inside the intake manifold. A low sensor reading indicates that incoming air is highly dense, while a high sensor reading indicates that the incoming air is extremely thin. To balance the proper air-to-fuel ratio, the Powertrain Control Modules (PCM) increase the amount of fuel to the engine in the event of a low-temperature reading, whereas the PCM reduces the amount of fuel to the engine in the event of a high-temperature reading.

\begin{figure*}[h]
    \centering
    \includegraphics[scale=0.65]{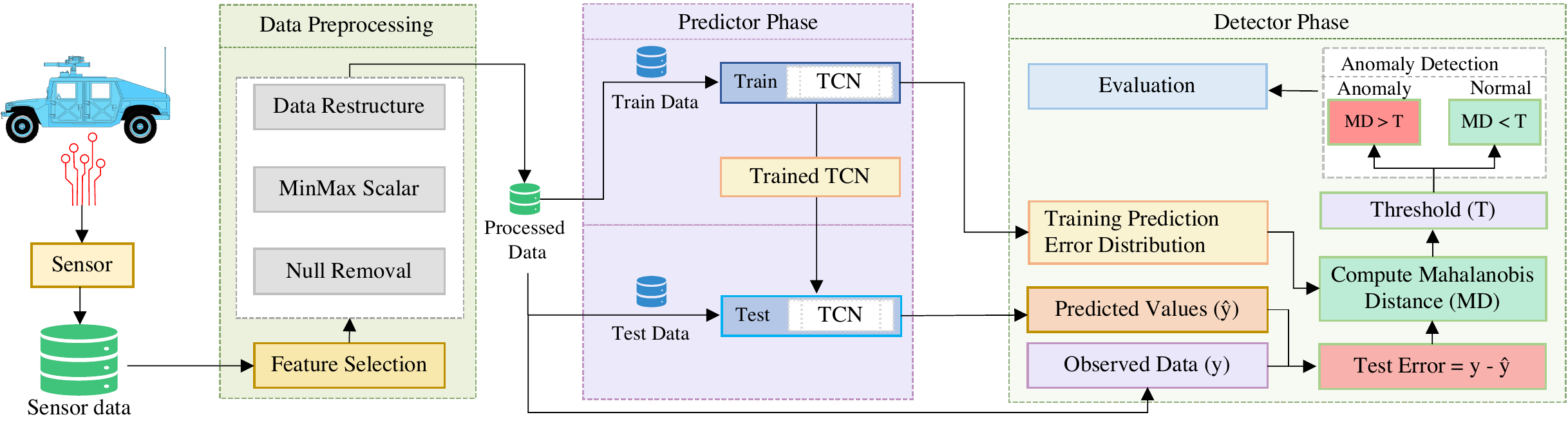}
    \caption{An overview of the proposed three-phased system architecture. In the initial phase, raw sensor data is cleansed, transformed, and scaled. The second phase predicts the sequence of outputs. In the third phase, anomalies are identified by comparing error values to a predefined threshold.}
        \label{fig: system_architecture}
\end{figure*} 
\subsection{Transmission FWG and Channels}
The transmission FWG controls and transfers power from the engine to the driveshaft. {The following sensor channels are associated with this FWG:}

\subsubsection{VehSpeedEng (Vehicle Speed)}
The \emph{VehSpeedEng} sensor channel represents the speed of a vehicle at a given time. 
The PCM manipulates numerous electrical subsystems in a vehicle using the \emph{VehSpeedEng} frequency signal, including fuel injection, ignition, cruise control operation, torque, and clutch lock-up. On the other hand, transmissions rely on vehicle speed data to optimize shift strategy, and it is a significant determinant of fuel usage.

\subsubsection{TransOilTemp (Transmission Oil Temperature)}
The \emph{TransOilTemp} sensor channel captures the temperature of the transmission fluid under all operating conditions. It measures transmission fluid temperature by transmitting a voltage signal proportional to fluid temperature to the ECU; i.e., the lower the voltage, the higher the temperature.

\subsubsection{TrSelGr (Transmission Selected Gear)}
The \emph{TrSelGr} sensor channel represents what transmission gear the vehicle is currently in. The gear ratio values set a torque limit on the engine output for a certain range of transmission gears. 

\subsubsection{TransTorqConvLockupEngaged (Transmission Torque Converter Lockup Engaged)}

The \emph{TransTorqConvLockupEngaged} sensor channel indicates whether or not the torque converter lockup is engaged. It is represented in binary form, i.e., 0 for disengaged and 1 for engaged. The transmission lockup torque converter has a clutch. The engagement of this clutch locks the engine to the input shaft of the transmission, resulting in a direct 1:1 drive ratio. It is used to enhance fuel consumption efficiency.

\subsubsection{TrOutShaftSp (Transmission Output Shaft Speed)}

\emph{TrOutShaftSp} sensor channel is used to calculate the transmission gear ratio when in use. It detects the rotational speed of the output shaft, which is derived from the output shaft tone wheel.
The Transmission Electronic Control Unit (TECU) looks at the speeds of the input shaft, the output shaft, and the main shaft to figure out the gear ratios of the main case, the auxiliary case, and the whole transmission.

\subsection{Fuel FWG and Channels}
The fuel FWG is responsible for the fuel material needed to produce heat when ignited with oxygen. This FWG includes the following sensor channels:

\subsubsection{\emph{FuelRate} (Fuel Rate)}
The \emph{FuelRate} sensor channel essentially captures how quickly the vehicle is burning fuel. It measures the amount of fuel a vehicle consumes to go a specific distance \cite{national2011assessment}.

\subsubsection{InstFuelEco (Instantaneous Fuel Economy)}

The \emph{InstFuelEco} sensor channel represents current fuel economy at current vehicle velocity.

\subsubsection{InjCtlPres (Injector Control Pressure)}
The \emph{InjCtlPres} sensor channel is the health indicator of the injectors. It detects the pressure of the fuel entering the injectors. It checks the pressure of the fuel going into the injectors and sends that information to the computer so that it can make the changes needed for the best performance and efficiency.
 
\subsection{Brake FWG and Channel}
The brake FWG is responsible with inhibiting motion by absorbing energy.
The position of the brake pedal is captured by the \emph{BrakeSwitch} sensor channel. The brake switch's sensor measures this position. Its principal purpose is to release the converter clutch during deceleration.

\section{System Architecture \& Methodology}
\label{architecture}

For a vehicle to work normally, it is important that all of its Functional Working Groups (FWG) work properly.  When one of the vehicle's FWG fails, the vehicle's performance deteriorates. Examples of FWGs include engine, fuel, cooling, air compressor, electrical, transmission, axle, suspension, brake, wheel, and tire related channels. For this study, four FWGs: engine, transmission, fuel, and brakes, have been included for modeling and evaluation (Figure \ref{fig:vechicle_subsystems_diagram}). 

The architecture of our system, shown in Figure \ref{fig: system_architecture}, is divided {into} three phases: data preprocessing, the prediction phase, and the anomaly detection phase. We {developed} our anomaly detection methodology utilizing this phased approach. In the rest of this section, we describe each phase and expand on our methods.

\subsection{Data Preprocessing}

\begin{figure}[h]
\centering
\includegraphics[width=0.45\textwidth]{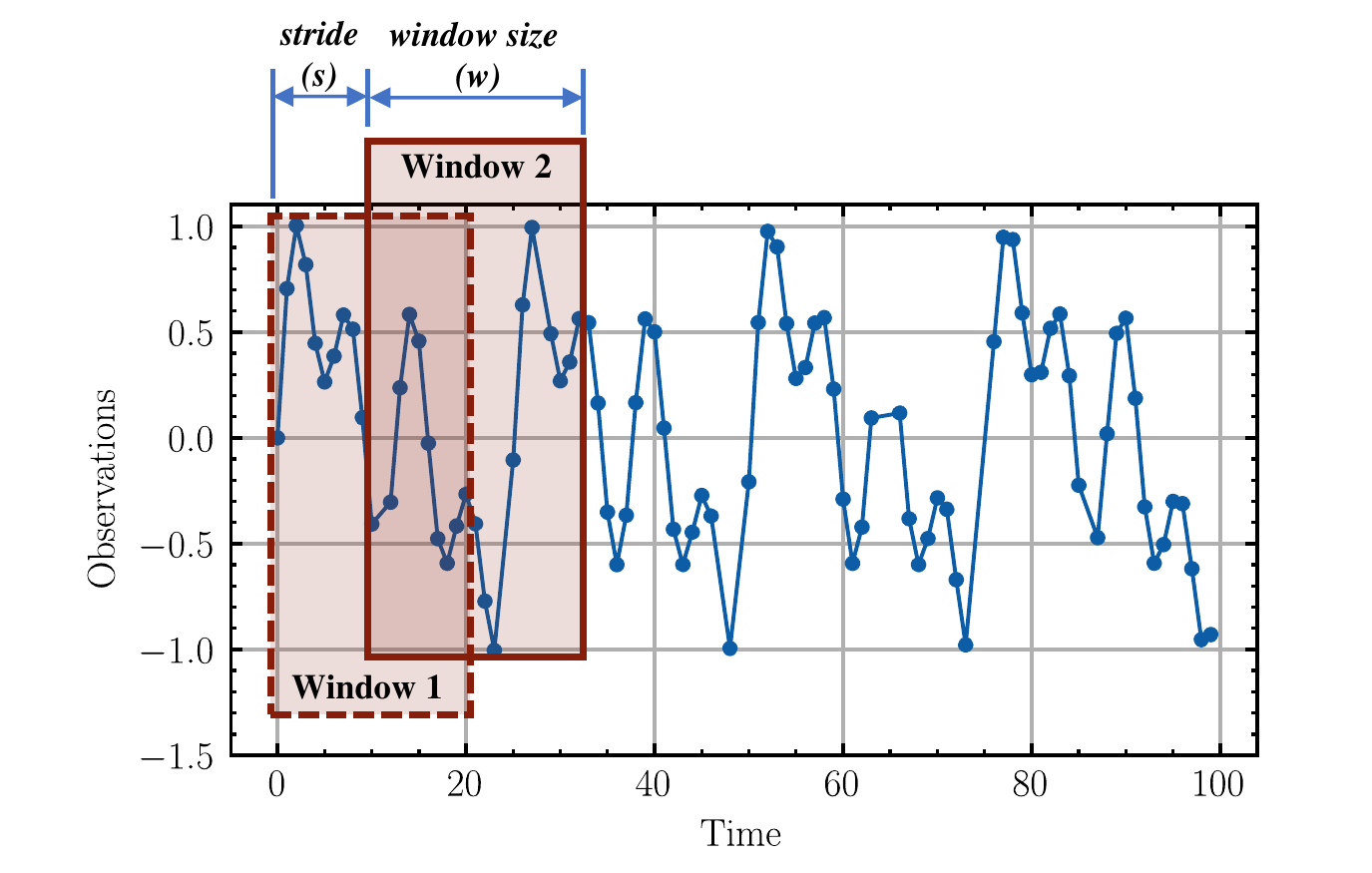}
\caption{Sliding window example. A window is comprised of $w$ = 20 observations. Consecutive windows can be separated by utilizing a stride value.}
\label{fig:sliding_window}
\vspace{-4mm}
\end{figure}

In the first phase of our architecture, the multi-channel data stream is preprocessed before the predictor phase. The dataset used in this study has 106 sensor channels with 955,856 observations. We selected 15 suitable sensor channels based on the advice of mechanical engineers working on the same data. These have been described in Section \ref{dataset_des}. Figure \ref{fig:sequence_example} shows raw data observations for these sensor channels during a window of 25 minutes collected on 30-Jan-2014. To process the data, we removed null values and redundant observations from the dataset. Next, we normalized the data using the min-max scaling. The dataset is further processed using a sliding window approach and conditioned to meet the input requirements for the next phase (See Figure \ref{fig:sliding_window} as an example). At the end of this phase, for a given time window, the processed data will be formatted as a sequence containing a certain number of observations per channel.

\subsection{Predictor Phase}

In the second phase, the predictor performs time series forecasting. TCN is utilized as the predictor model due to its ability to abstract time series data and its use of flexible receptive fields \cite{he2019temporal}. In addition, these models are computationally efficient and possess representational capability to achieve robust and superior prediction performance \cite{bai2018empirical}. In our model, time series forecasting is defined as predicting the next value from an input sequence of observations.  

A sequence-to-sequence (Seq2Seq) learning approach is used due to the temporal nature of the dataset. 
The processed data is restructured as sequences, i.e., series of contiguous observations. In order to generate sequences, a sliding window is used on the processed data. Figure \ref{fig:sliding_window}, illustrates an example of the sliding window approach. The size of the \emph{window (w)} is the number of observations to consider as a sequence. In order to avoid interpolation and to ensure that sequences are composed of contiguous observations, the \emph{time elapsed $(et)$} between the first and last observation in a window (endpoints) is calculated. The \emph{expected data rate $(dr)$} for this dataset is 1 Hertz (Hz). Due to dropped packets in streaming data, an additional \emph{delta time $(dt)$} is added when considering sequences.  For example, consider a sequence of 10 observations with a 1 Hz data rate that has an elapsed time of 12 seconds. This sequence would be considered valid if $dt$ is set to 2. In other words, a sequence is valid if: $et \leq (w/dr + dt)$ seconds. All other sequences are removed since they contain non-contiguous observations. Non-contiguous observations can be due to gaps in the data when the vehicle is idle or not in use. A fixed stride is also used to control the amount of overlap between windows. 
{The final processed dataset is a three-dimensional array of the form $N\times M\times P$, where \emph{N} is the number of instances, \emph{M} is the length of the sequence, and \emph{P} is the number of channels.} For our experiments, the restructured dataset has 15 sensor channels, each with $w$ = 20 observations.

The restructured data is split into two parts: training and testing. For this study, we used 70\%–30\% random splitting criteria for training and testing. Due to the random split, training and testing datasets contain sequences throughout the 21 month period between 2012 and 2014. Each sequence is broken into two sub-sequences. 
The first $M - 1$ observations are used as the input sequence, and the last one is used as the target output.
The predictor is a sequential model with an temporal convolutional layer and a dense layer. The temporal convolutional layer has the following hyperparameters: 64 filters; kernel size of 3; causal padding; single residual block stack; ReLU activation function; and dilations of 1, 2, 4, 8, 16, and 32. The size of the receptive field is 505 (See Section \ref{tcnback} for details). The dense layer has $1 \times P$ units and uses a linear activation function. A 10\% validation split is used on the training data to ensure that the model does not overfit. All the {abovementioned} hyperparameters were determined experimentally during optimization. 

\input{testtable}
\input{test_results_figures}

\input{algo}
\subsection{Detector Phase}

In the final phase{,} we perform anomaly detection. The objective of this phase is to determine whether or not a FWG is functioning normally. The predicted observation, calculated in the previous phase, is compared to the actual observation. In general, if the prediction error exceeds a threshold, then the system estimates an abnormal observation. 

{The anomaly detection technique used in this paper is outlined in the Algorithm }\ref{algorithm}. Once the predictor has been trained, the training dataset is used to calculate the estimated prediction error. The estimated (training) prediction error is assumed to follow a multivariate Gaussian distribution. It is important to note that the training data may contain a very small number $(\alpha)$ of abnormal instances due to sensor measurement noise. Another {assumption} is made that the vehicle operates under normal conditions for a large number $(\beta)$ of training instances $($i.e., $\beta >> \alpha)$. The Mahalanobis distance is used to determine how closely the test prediction error matches the estimated distribution.

Anomalies are inserted in 20\% of the test dataset. Then, the predictor is used to make predictions on the test dataset. As stated before, the Mahalanobis distance is used as a metric to detect anomalies. Finally, the performance of the anomaly detection system is evaluated using receiver operating characteristic (ROC) curve and area under the curve (AUC) \ref{fig:roc_curve}. The MD threshold is derived from the ROC curve by using the geometric mean of sensitivity and specificity. The Confusion Matrix \ref{confusion_matrix} and detection accuracy can be formulated using a MD threshold.

\section{Evaluation and Results}
\label{evaluation}

To evaluate our system, we must ensure that the anomaly detection model is robust enough to predict both normal and abnormal states for FWGs. This means that when the sensor channels have normal readings, the model must not raise a flag. Conversely, if there is any abnormal data collected from the sensor channels, then the model should raise an alert. To test the system's ability to find sensor channels in states that are not normal, we mix in some synthetic data with real data to create a number of test scenarios. Next, we describe these anomalous text scenarios. These were also created in consultation with mechanical engineers working on the same dataset.

\subsection{Anomalous Test Scenarios}
To evaluate the vehicle's fuel system, we must consider three types of information: (1) performance sensor channels such as {\emph{FuelRate}}; (2) sensor channels that affects fuel rate such as \emph{AccelPedalPos, VehSpeedEng, PctEngLoad}, and \emph{EngPctTorq}; and (3) fuel rate associated sensor channels such as \emph{InjCtlPres, IntManfTemp}, and \emph{TrSelGr}. The detailed description of these channels can be found in Section \ref{dataset_des} and Table \ref{table:features_description}. On the basis of this information, we develop three scenarios containing 21 test conditions, as shown in Table \ref{table:test_conditions} . Our model should identify these anomalies accurately. For example, the model should not predict the first test case in Table \ref{table:a}  as an anomaly, whereas it should accurately predict the second case as an anomaly.

\subsubsection{Scenario 1}
\label{result:scenario1}
When the \emph{AccelPedalPos} is pushed, the ECU opens the throttle body to increase the flow of fuel to the engine and closes it when the pedal is released. Ideally, with rapid acceleration, the \emph{TrSelGr} should be higher. Using this scenario, we construct seven test cases as depicted in Table \ref{table:a}.

\subsubsection{Scenario 2}
\label{result:scenario2}

In the normal operating conditions, if the \emph{IntManfTemp} sensor channel experiences a low reading, then it is considered that the air coming to the intake manifold is highly dense, and if the reading is high, the air is considered to be thin. To balance things, more fuel is injected into the engine when the internal manifold temperature is low, and fuel flow is reduced when the manifold temperature is high. Using this scenario, we generate the seven test cases as shown in Table \ref{table:b}. 

\subsubsection{Scenario 3}
\label{result:scenario3}
In the case of the \emph{InjCtlPres} sensor channel, when the pressure is increased in the injectors, it causes the valve to open wide, leading to an increase in fuel flow into the engine. Based on this assumption, the test conditions shown in Table \ref{table:c} are formulated.

\subsection{Insertion of Anomalous Observations to Test Dataset}

Using the abovementioned test scenarios, anomalous observations are inserted into the testing data. Figure \ref{FIGURE:LABEL} illustrates an example of inserting synthetic anomalies into the test data and predictions made by the TCN model. This is done by assigning 20\% of testing instances an anomaly scenario (type) at random. The assigned scenario dictates the particular sensor channels and the direction of modification. If an instance is chosen to be anomalous, then the target observation is modified based of the assigned scenario. The magnitude of the modification is also chosen at random from the set (1, 1.5). Note that modified values are bounded between -1 and 1.

\subsection{Results}
Before discussing the results, we define performance metrics that we used in the context of our anomaly detection system. We classify a sensor channel reading as a true positive (TP) only if the model detects an abnormal sensor channel reading as an anomaly, and a true negative (TN) when a normal channel sensor reading is detected as normal. When the model detects a normal sensor channel reading as an anomaly, it is defined as a false
positive (FP), whereas an actual abnormal sensor channel reading which is not detected is a false negative (FN). Using these definitions, we evaluate the proposed method based on two performance metrics: 

\begin{enumerate}
    \item Detection Accuracy: Quantifies the detection system's ability to correctly identify an anomaly, as defined below: 
    \begin{equation}
\text { Detection accuracy }=\frac{T P+T N}{T P+F P+T N+F N}
\end{equation}
    
    \item Receiver Operating Characteristic (ROC) curve with area under the curve (AUC): Measures the performance of the proposed model as the area under the curve in a plot between the true positive rate (TPR) and false positive rate (FPR):
    \begin{equation}
T P R=\frac{T P}{T P+F N}\quad, \ \ F P R=\frac{F P}{F P+T N}
\end{equation}
\end{enumerate}

\begin{figure}[h]
\centering
\includegraphics[width=0.35\textwidth]{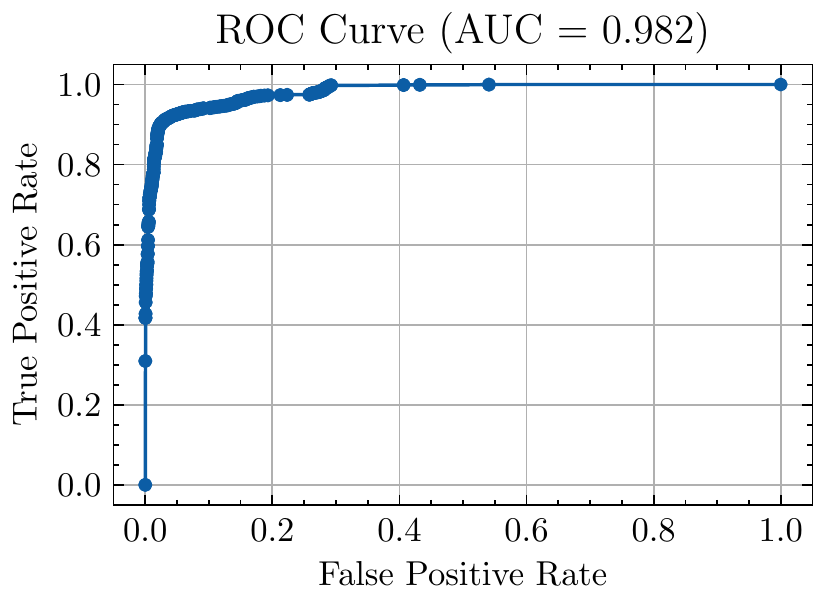}
\caption{ROC Curve on test dataset.}
\label{fig:roc_curve}
\vspace{-4mm}
\end{figure}

The predictor TCN model was trained using validation-based early stopping. The overall result of the anomaly detection system is shown using the ROC curve (see Figure \ref{fig:roc_curve}). The area under the curve (AUC) of the system is 0.982. Using a MD threshold 9.61, 91\% of true anomalies were detected. The proposed system has an detection accuracy of 0.96 (See Table \ref{confusion_matrix}).

\begin{table}[h]
\centering
{\renewcommand{\arraystretch}{1.3}%
\begin{tabular}{l|l|c|c|c}
\multicolumn{2}{c}{}&\multicolumn{2}{c}{Predicted}&\\
\cline{3-4}
\multicolumn{2}{c|}{}&Normal&Anomaly\\
\cline{2-4}
\multirow{2}{*}{\rotatebox[origin=c]{90}{Actual}}& Normal & \cellcolor{green!15}$6942 \ (97\%)$ & $221 \ (3\%)$ \\
\cline{2-4}
& Anomaly & $158 \ (9\%)$ & \cellcolor{green!25}$1632 \ (91 \%)$ \\
\cline{2-4}

\end{tabular}}
 \caption{Confusion matrix. A Mahalanobis distance threshold of 9.61 is used to detect anomalies.}
 \label{confusion_matrix}
\end{table}

\input{confusion_matrix_table}

The composition of anomalous instances per scenario was: 518 from Scenario 1, 707 from Scenario 2, and 565 from Scenario 3. In total, there were 1790 anomalous instances. The following is the breakdown of missed anomalies by the system: 49 missed in Scenario 1, 3 missed in Scenario 2, and 106 missed in Scenario 3. This suggests that the system performs better when testing with channels from multiple functional working groups. Additionally, by using the TCN model prediction, the system can further isolate the anomaly to a specific channel and FWG (see Figure \ref{SUBFIGURE LABEL 2}). Overall, the system successfully identified vehicular anomalies and states well outside  normal operating conditions (e.g., potentially unsafe states). 

\section{Conclusion \& Future Work}
\label{conclusion}
In this paper, we introduce a comprehensive anomaly detection system for vehicles utilizing multiple Functional Working Groups (FWGs). FWGs provide information to the anomaly detection system in the form of multisensor time series data. The anomaly detection model was developed using the Vehicle Performance, Reliability and Operations (VePRO) dataset, which contains usage data from a fleet of U.S. Department of Defense vehicles.   

To detect anomalies, a multi-phased approach is employed with a Temporal Convolution Network (TCN) used as a kernel to learn the underlying relationships between sensor channels. The TCN model was trained using the VePRO data, which contains normal operating conditions. The model performs time series forecasting by operating on a multi-channel sequence for the current time window and outputs a prediction for the next observation. The final anomaly determination is made by comparing the prediction with the actual observation. Through a series of targeted scenarios and tests, the system is able to isolate abnormal behavior to a specific FWG. 
The results of this study show that our system successfully identifies vehicular anomalies and unsafe states.

In the future, a logical next step is to expand on the forecasting model to consider long sequences using RNN-based LSTM methods. Given the architectural differences between LSTM and TCN methods, it would be interesting to compare the predictive and anomaly detection capabilities of these two algorithms. In addition, we would like to investigate the performance of hybrid approaches, such as ConvLSTM \cite{shi2015convolutional}, for anomaly detection in vehicular operational datasets. 

\section*{Acknowledgement}
This Work was supported by PATENT Lab (Predictive Analytics and TEchnology iNTegration Laboratory) at the Department of Computer Science and Engineering, Mississippi State University; U.S. Department of Defense grant through U.S. Army Engineer Research and Development Center (ERDC); National Science Foundation (\#1565484). The views and conclusions are those of the authors.

\bibliographystyle{unsrt}
\bibliography{refs}

\end{document}

%% file: Table/channel_summary.tex
\begin{table*}[ht]
{\renewcommand{\arraystretch}{1.20}%
\begin{tabular}{|p{1.5cm}|p{2cm}|p{2.3cm}|p{2.7cm}|p{7.3cm}|}
\hline
\rowcolor{lightgray!20!}
FWG         & Sensor Channel                    & Full name                                     & Unit                           & Description                                                                        \\ \hline
Time         & UTC\_1HZ                   & 1 Hz Data                                    & Seconds                            & Time series data from the 1 Hz channels.                                       \\ \hline

                                    & EngCoolantTemp             & Engine Coolant Temperature                   & Degree Fahrenheit (°F)            & Captures how heavy the engine is working and correlates well   with the fuel rate. \\\cline{2-5}  
                                    & PctEngLoad                 & Percentage Engine Load                       & Percent (\%)                             & Amount of load required for the engine to perform driving.                         \\\cline{2-5}  
\multirow{6}{*}{Engine}              & EngPctTorq                 & Engine Percentage Torque                     & Percent (\%)                            & The calculated output torque of the engine.                           \\\cline{2-5}  
                                    & BoostPres                  & Turbo Boost Pressure                         & Pound per square inch (psi)              & Indicates the air pressure information at any given moment.                        \\\cline{2-5}  
                                     & AccelPedalPos              & Acclerator Pedal Position                    & Percent (\%)                             & Captures acceleration, deceleration, and steady state   condition.                 \\\cline{2-5}  
                                    & IntManfTemp                & Intake Manifold Temperature                  & Degree Fahrenheit (°F)                  & Indicates the temperature of the air inside the intake   manifold.                 \\\cline{2-5}  \hline
                                    

& VehSpeedEng                & Vehicle Speed                                & Miles Per Hour (m/s)                   & Speed of vehicle -  major contributor to the fuel consumption.                      \\\cline{2-5} 
             & TransOilTemp               & Transmission Oil Temperature                 & Degree Fahrenheit (°F)                   & Captures temperature in all operating condition.                                   \\\cline{2-5} 
\multirow{6}{*} {Transmission} & TrSelGr                    & Transmission Selected Gear                   & Unitless                                 & Represents which transmission gear the vehicle is currently in.                  \\\cline{2-5}  
             & TransTorqConv
             LockupEngaged & Transmission Torque Converter Lockup Engaged & Base 10 Integer Number                  & Captures whether or not the torque converter lockup is   engaged.                 \\\cline{2-5}  
             & TrOutShaftSp               & Transmission Output Shaft Speed              & Revolutions Per Minute (rpm)            & Calculates the transmission gear ratio when in use.                                \\\cline{2-5}  \hline

 & FuelRate                   & Fuel Rate                                    & Gallons (U.S.) Per Hour (gph)             & The fuel rate is essentially how   quickly the vehicle is burning fuel.             \\\cline{2-5} 
\multirow{3}{*} {Fuel}         & InstFuelEco                & Instantaneous Fuel Economy                   & Gallons (U.S.) Per Hour (gph)             & Captures current fuel economy at current vehicle velocity.                           \\\cline{2-5}
             & InjCtlPres                 & Injector Control Pressure                    & Pound per square inch (psi)               & Can be used as a health indicator of the injectors.                                 \\\cline{2-5} \hline


\multirow{1}{*} {Brake}        & BrakeSwitch                & Brake Switch                                 & Base 10 Integer Number                  & Captures the position of break pedal.        \\\cline{2-5}     \hline     

                                    
\end{tabular}}
\\
\\
\caption{A brief overview of the various vehicular FWGs included in the study, along with a description of the associated sensor channels.}

\label{table:features_description}

\vspace{-4mm}

\end{table*}

%% file: testtable.tex
\begin{table*}

            \footnotesize
{\renewcommand{\arraystretch}{1.20}%
\subcaptionbox{Scenario 1 \label{table:a}}{
\begin{tabular}{r|l|l|l|c}
\rowcolor{lightgray!20!}
TC & FuelRate & AccelPedalPos & TrSelGr & \multicolumn{1}{l}{Result}  \\ \hline
1                       & $\Leftrightarrow $        & $\Leftrightarrow $            & $\Leftrightarrow $       & FALSE                       \\

2                       & $\Leftrightarrow $       & $\Downarrow$             & $\Downarrow$      & TRUE                        \\

3                       & $\Leftrightarrow $       & $\Uparrow$             & $\Downarrow$     & TRUE                        \\

4                       & $\Uparrow$        & $\Uparrow$            & $\Uparrow$      & FALSE                       \\

5                       & $\Uparrow$      & $\Downarrow$              & $\Downarrow$         & TRUE                        \\

6                       & $\Downarrow$         & $\Uparrow$             & $\Uparrow$       & TRUE                        \\

7                      & $\Downarrow$        & $\Downarrow$              & $\Downarrow$        & FALSE                      
\end{tabular}\label{table:two_a}}
}
                \hfill
                {\renewcommand{\arraystretch}{1.20}%
                \subcaptionbox{Scenario 2 \label{table:b}}{
\begin{tabular}{r|l|l|c}
\rowcolor{lightgray!20!}
TC & FuelRate & IntManfTemp & \multicolumn{1}{l}{Result}  \\\hline
1                       & $\Leftrightarrow$         & $\Leftrightarrow $            & FALSE                       \\
2                       & $\Leftrightarrow $         & $\Uparrow$           & TRUE                        \\
3                       & $\Leftrightarrow $        & $\Downarrow$           & TRUE                        \\
4                       & $\Uparrow$        & $\Downarrow$            & FALSE                       \\
5                       & $\Downarrow$         & $\Uparrow$            & TRUE                        \\
6                       & $\Uparrow$        & $\Uparrow$          & TRUE                        \\
7                       & $\Downarrow$         & $\Downarrow$           & TRUE                       
\end{tabular}}}
                \hfill
                {\renewcommand{\arraystretch}{1.20}%
                \subcaptionbox{Scenario 3 \label{table:c}}{
                    \begin{tabular}{r|l|l|c}
                    \rowcolor{lightgray!20!}
                        TC & FuelRate & InjCtlPres & \multicolumn{1}{l}{Result}  \\ \hline
1                       & $\Leftrightarrow$         & $\Leftrightarrow$          & FALSE                       \\
2                       & $\Leftrightarrow$        & $\Uparrow$          & TRUE                        \\
3                       & $\Leftrightarrow$        & $\Downarrow$          & TRUE                        \\
4                       & $\Uparrow$       & $\Downarrow$          & TRUE                        \\
5                       & $\Downarrow$         & $\Uparrow$         & TRUE                        \\
6                       & $\Uparrow$        & $\Uparrow$         & FALSE                       \\
7                       & $\Downarrow$         & $\Downarrow$           & FALSE                      
\end{tabular}}}

                \caption{$\Leftrightarrow $ represent normal condition, $\Uparrow$  represent rise and $\Downarrow$  represent fall. TC refers to test cases for a scenario. Scenario 1 illustrates the normal and abnormal relationships between the \emph{FuelRate, AccelPedalPos}, and \emph{TrSelGr} channels. The second scenario presents the relationship between \emph{FuelRate} and \emph{IntManfTemp} in normal and abnormal conditions. Scenario 3 illustrates the normal and abnormal relationship between \emph{FuelRate} and \emph{InjCtrlPres}.}
                \label{table:test_conditions}
            \end{table*}

%% file: test_results_figures.tex
\begin{figure*}
\captionsetup[subfigure]{justification=centering}
\centering  
\begin{subfigure}{.36\textwidth}
    \centering
    \includegraphics[width=\linewidth]{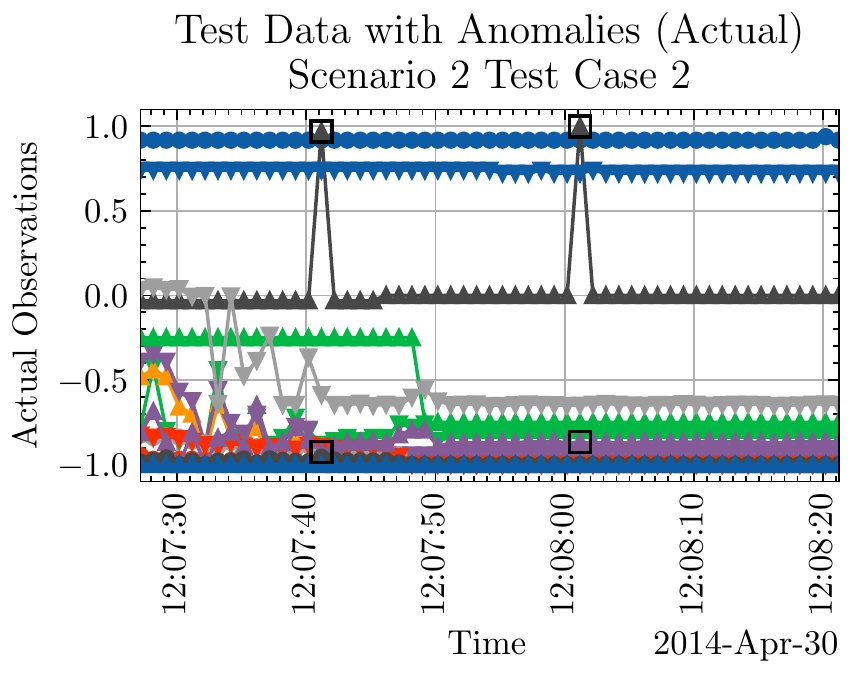}
     \caption{Synthetic anomalies inserted into test data.}
    \label{SUBFIGURE LABEL 1}
\end{subfigure}
\begin{subfigure}{.36\textwidth}
    \centering
    \includegraphics[width=\linewidth]{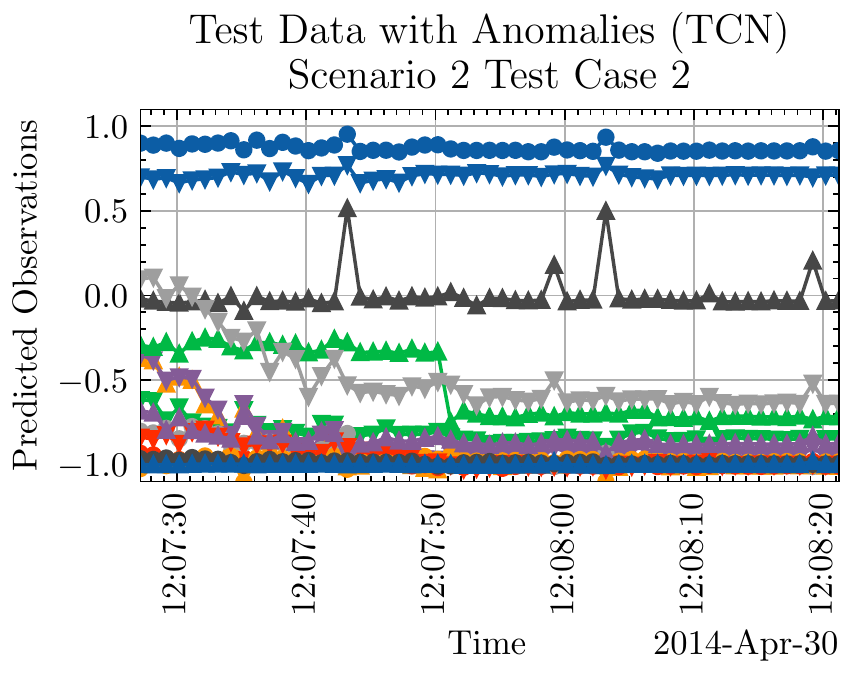}
     \caption{TCN predictions on test data.}
    \label{SUBFIGURE LABEL 2}
\end{subfigure}
\begin{subfigure}{.25\textwidth}
    \centering
    \includegraphics[width=\linewidth]{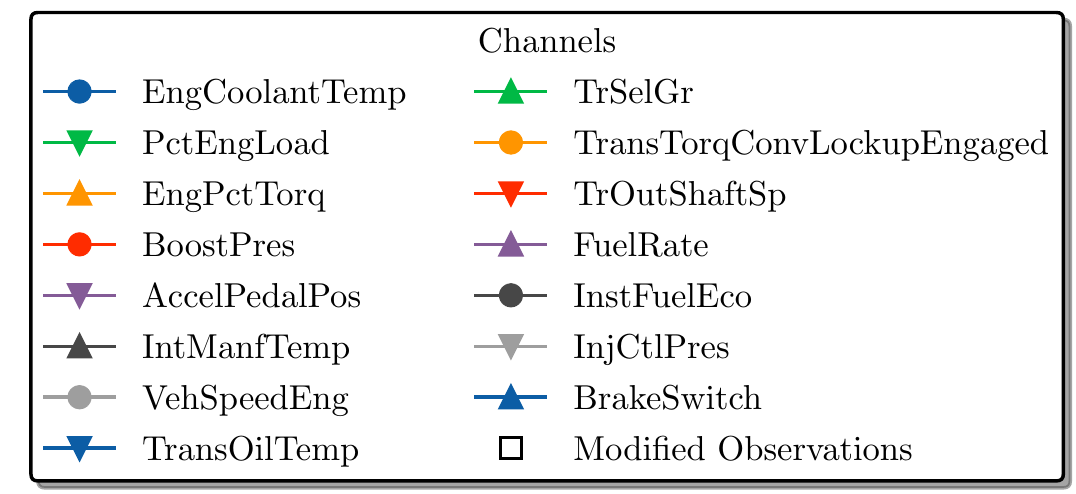}  
    \label{SUBFIGURE LABEL 3}
\end{subfigure}

\caption{Plot (a) illustrates a snapshot of synthetic anomalies mixed into the test data for a time period of 54 seconds on 2014-Apr-30. Scenario 2, Test Case 2 involves an increase in \emph{IntManfTemp} and an unchanged \emph{FuelRate}. Modifications are bounded within -1.0 and 1.0. Black square markers denote modified observations. Plot (b) shows TCN predictions on the test data.}
\label{FIGURE:LABEL}
\vspace{-4mm}
\end{figure*}

%% file: algo.tex
\newcommand{\algrule}[1][.2pt]{\par\vskip.5\baselineskip\hrule height #1\par\vskip.5\baselineskip}

\begin{algorithm}
\caption{Training and Testing a TCN based Anomaly Detection System}
\label{algorithm}
\textbf{{Inputs}}:
\\ \emph{$data$} multivariate time series observations
\\ \emph{$tcn_{hp}$} TCN hyperparameters
\\ \textbf{{Outputs}}:
\\ \emph{$Model$} trained temporal convolutional network model
\\ \emph{$\mu_{pred}$} training prediction error mean
\\ \emph{$\Sigma_{pred}$} training prediction error covariance
\\ \emph{$T$} anomaly threshold
\\ \emph{$CM$} confusion matrix
\\ \textbf{{Variables}}:
\\ \emph{$X$} model input, \emph{$y$} model target, \emph{$\hat{y}$} model prediction, \emph{$fa$}  actual anomaly label, \emph{$fp$} predicted anomaly label, \emph{$trn$} training data, \emph{$tst$} test data, \emph{$\zeta$} prediction error, \emph{$MD$} Mahalanobis distance
 \algrule
\begin{algorithmic}[1]
\State $ X_{trn}, y_{trn}, X_{tst}, y_{tst}, fa_{tst}= \ $split\_dataset$ $$(data) $
\State $Model \gets \ $build\_model$(tcn_{hp}, X\_trn, y\_trn)$
\State $\hat{y}_{trn} = \ $predict$(\emph{$Model$}, X_{trn})$
\State $\zeta_{trn} = y_{trn_i} - \hat{y}_{trn_i}$
\State $\mu_{pred} = \ $mean$(\zeta_{trn})$
\State $\Sigma_{pred} = \ $cov$(\zeta_{trn})$
 \algrule
\State $\hat{y}_{tst} = \ $predict$(\emph{Model}, X_{tst})$
\State $\zeta_{tst} = y_{tst} - \hat{y}_{tst}$
\State Select $T$
\State $\emph{$MD$} = \sqrt{(\zeta_{tst} - \mu_{pred})^{T} \ \Sigma_{pred}^{-1} \ (\zeta_{tst} - \mu_{pred})}$
 \If{$MD > \textit{T}$}
    \State $fp_{tst} = \emph{Anomaly}$
\Else
   \State $fp_{tst} = \emph{Normal}$
\EndIf 
\State $CM = confusion\_matrix(fp_{tst}, fa_{tst})$

\end{algorithmic}
\end{algorithm}

%% file: conference_041818.bbl
\begin{thebibliography}{10}

\bibitem{ran2019survey}
Yongyi Ran, Xin Zhou, Pengfeng Lin, Yonggang Wen, and Ruilong Deng.
\newblock A survey of predictive maintenance: Systems, purposes and approaches.
\newblock {\em arXiv preprint arXiv:1912.07383}, 2019.

\bibitem{zhang2019data}
Weiting Zhang, Dong Yang, and Hongchao Wang.
\newblock Data-driven methods for predictive maintenance of industrial
  equipment: A survey.
\newblock {\em IEEE Systems Journal}, 13(3):2213--2227, 2019.

\bibitem{zonta2020predictive}
Tiago Zonta, Cristiano~Andr{\'e} Da~Costa, Rodrigo da~Rosa~Righi, Miromar~Jose
  de~Lima, Eduardo~Silveira da~Trindade, and Guann~Pyng Li.
\newblock Predictive maintenance in the industry 4.0: A systematic literature
  review.
\newblock {\em Computers \& Industrial Engineering}, 150:106889, 2020.

\bibitem{carvalho2019systematic}
Thyago~P Carvalho, Fabr{\'\i}zzio~AAMN Soares, Roberto Vita, Roberto da~P
  Francisco, Jo{\~a}o~P Basto, and Symone~GS Alcal{\'a}.
\newblock A systematic literature review of machine learning methods applied to
  predictive maintenance.
\newblock {\em Computers \& Industrial Engineering}, 137:106024, 2019.

\bibitem{chalapathy2019deep}
Raghavendra Chalapathy and Sanjay Chawla.
\newblock Deep learning for anomaly detection: A survey.
\newblock {\em arXiv preprint arXiv:1901.03407}, 2019.

\bibitem{kwon2019survey}
Donghwoon Kwon, Hyunjoo Kim, Jinoh Kim, Sang~C Suh, Ikkyun Kim, and Kuinam~J
  Kim.
\newblock A survey of deep learning-based network anomaly detection.
\newblock {\em Cluster Computing}, 22(1):949--961, 2019.

\bibitem{ma2021comprehensive}
Xiaoxiao Ma, Jia Wu, Shan Xue, Jian Yang, Chuan Zhou, Quan~Z Sheng, Hui Xiong,
  and Leman Akoglu.
\newblock A comprehensive survey on graph anomaly detection with deep learning.
\newblock {\em IEEE Transactions on Knowledge and Data Engineering}, 2021.

\bibitem{chandola2009anomaly}
Varun Chandola, Arindam Banerjee, and Vipin Kumar.
\newblock Anomaly detection: A survey.
\newblock {\em ACM computing surveys (CSUR)}, 41(3):1--58, 2009.

\bibitem{alizadeh2021vehicle}
Morteza Alizadeh, Michael Hamilton, Parker Jones, Junfeng Ma, and Raed Jaradat.
\newblock Vehicle operating state anomaly detection and results virtual reality
  interpretation.
\newblock {\em Expert Systems with Applications}, 177:114928, 2021.

\bibitem{he2019temporal}
Yangdong He and Jiabao Zhao.
\newblock Temporal convolutional networks for anomaly detection in time series.
\newblock In {\em Journal of Physics: Conference Series}, volume 1213, page
  042050. IOP Publishing, 2019.

\bibitem{bai2018empirical}
Shaojie Bai, J~Zico Kolter, and Vladlen Koltun.
\newblock An empirical evaluation of generic convolutional and recurrent
  networks for sequence modeling.
\newblock {\em arXiv preprint arXiv:1803.01271}, 2018.

\bibitem{lea2017temporal}
Colin Lea, Michael~D Flynn, Rene Vidal, Austin Reiter, and Gregory~D Hager.
\newblock Temporal convolutional networks for action segmentation and
  detection.
\newblock In {\em proceedings of the IEEE Conference on Computer Vision and
  Pattern Recognition}, pages 156--165, 2017.

\bibitem{long2015fully}
Jonathan Long, Evan Shelhamer, and Trevor Darrell.
\newblock Fully convolutional networks for semantic segmentation.
\newblock In {\em Proceedings of the IEEE conference on computer vision and
  pattern recognition}, pages 3431--3440, 2015.

\bibitem{alla2019beginning}
Sridhar Alla and Suman~Kalyan Adari.
\newblock {\em Beginning anomaly detection using python-based deep learning}.
\newblock Springer, 2019.

\bibitem{oord2016wavenet}
Aaron van~den Oord, Sander Dieleman, Heiga Zen, Karen Simonyan, Oriol Vinyals,
  Alex Graves, Nal Kalchbrenner, Andrew Senior, and Koray Kavukcuoglu.
\newblock Wavenet: A generative model for raw audio.
\newblock {\em arXiv preprint arXiv:1609.03499}, 2016.

\bibitem{yu2015multi}
Fisher Yu and Vladlen Koltun.
\newblock Multi-scale context aggregation by dilated convolutions.
\newblock {\em arXiv preprint arXiv:1511.07122}, 2015.

\bibitem{he2016deep}
Kaiming He, Xiangyu Zhang, Shaoqing Ren, and Jian Sun.
\newblock Deep residual learning for image recognition.
\newblock In {\em Proceedings of the IEEE conference on computer vision and
  pattern recognition}, pages 770--778, 2016.

\bibitem{CoverHMMWVTM}
{OPERATOR’S MANUAL} us army ground vehicles.
\newblock \url{https://www.hmmwvinscale.com/HMMWV\%20TM.pdf}, 1993.
\newblock Accessed: 06-20-2022.

\bibitem{liu2019anomaly}
Jianwei Liu, Hongwei Zhu, Yongxia Liu, Haobo Wu, Yunsheng Lan, and Xinyu Zhang.
\newblock Anomaly detection for time series using temporal convolutional
  networks and gaussian mixture model.
\newblock In {\em Journal of Physics: Conference Series}, volume 1187, page
  042111. IOP Publishing, 2019.

\bibitem{thiruloga2022tenet}
Sooryaa~Vignesh Thiruloga, Vipin~Kumar Kukkala, and Sudeep Pasricha.
\newblock Tenet: Temporal cnn with attention for anomaly detection in
  automotive cyber-physical systems.
\newblock In {\em 2022 27th Asia and South Pacific Design Automation Conference
  (ASP-DAC)}, pages 326--331. IEEE, 2022.

\bibitem{gopali2021comparison}
Saroj Gopali, Faranak Abri, Sima Siami-Namini, and Akbar~Siami Namin.
\newblock A comparison of tcn and lstm models in detecting anomalies in time
  series data.
\newblock In {\em 2021 IEEE International Conference on Big Data (Big Data)},
  pages 2415--2420. IEEE, 2021.

\bibitem{wang2021anomaly}
Yuan Wang, Yan Wu, Qiong Yang, and Jun Zhang.
\newblock Anomaly detection of spacecraft telemetry data using temporal
  convolution network.
\newblock In {\em 2021 IEEE International Instrumentation and Measurement
  Technology Conference (I2MTC)}, pages 1--5. IEEE, 2021.

\bibitem{mo2021unsupervised}
Ronghong Mo, Yiyang Pei, Neelakantam Venkatarayalu, Pereira Nathaniel,
  AB~Premkumar, and Sumei Sun.
\newblock An unsupervised tcn-based outlier detection for time series with
  seasonality and trend.
\newblock In {\em 2021 IEEE VTS 17th Asia Pacific Wireless Communications
  Symposium (APWCS)}, pages 1--5. IEEE, 2021.

\bibitem{alizadeh2022hybrid}
Morteza Alizadeh, Shahram Rahimi, and Junfeng Ma.
\newblock A hybrid arima--wnn approach to model vehicle operating behavior and
  detect unhealthy states.
\newblock {\em Expert Systems with Applications}, 194:116515, 2022.

\bibitem{9659838}
Somayeh~Bakhtiari Ramezani, Brad Killen, Logan Cummins, Shahram Rahimi, Amin
  Amirlatifi, and Maria Seale.
\newblock A survey of hmm-based algorithms in machinery fault prediction.
\newblock In {\em 2021 IEEE Symposium Series on Computational Intelligence
  (SSCI)}, pages 1--9, 2021.

\bibitem{narayanan2016obd_securealert}
Sandeep~Nair Narayanan, Sudip Mittal, and Anupam Joshi.
\newblock Obd\_securealert: An anomaly detection system for vehicles.
\newblock In {\em 2016 IEEE International Conference on Smart Computing
  (SMARTCOMP)}, pages 1--6. IEEE, 2016.

\bibitem{chukkapalli2021privacy}
Sai Sree~Laya Chukkapalli, Priyanka Ranade, Sudip Mittal, and Anupam Joshi.
\newblock A privacy preserving anomaly detection framework for cooperative
  smart farming ecosystem.
\newblock In {\em 2021 Third IEEE International Conference on Trust, Privacy
  and Security in Intelligent Systems and Applications (TPS-ISA)}, pages
  340--347. IEEE, 2021.

\bibitem{ramapatruni2019anomaly}
Sowmya Ramapatruni, Sandeep~Nair Narayanan, Sudip Mittal, Anupam Joshi, and
  Karuna Joshi.
\newblock Anomaly detection models for smart home security.
\newblock In {\em 2019 IEEE 5th Intl Conference on Big Data Security on Cloud
  (BigDataSecurity), IEEE Intl Conference on High Performance and Smart
  Computing,(HPSC) and IEEE Intl Conference on Intelligent Data and Security
  (IDS)}, pages 19--24. IEEE, 2019.

\bibitem{narayanan2016using}
Sandeep~Nair Narayanan, Sudip Mittal, and Anupam Joshi.
\newblock Using semantic technologies to mine vehicular context for security.
\newblock In {\em 2016 IEEE 37th Sarnoff Symposium}, pages 124--129. IEEE,
  2016.

\bibitem{nair2015using}
Sandeep Nair~Narayanan, Sudip Mittal, and Anupam Joshi.
\newblock Using data analytics to detect anomalous states in vehicles.
\newblock {\em arXiv e-prints}, pages arXiv--1512, 2015.

\bibitem{kang2021anomaly}
Jaeyong Kang, Chul-Su Kim, Jeong~Won Kang, and Jeonghwan Gwak.
\newblock Anomaly detection of the brake operating unit on metro vehicles using
  a one-class lstm autoencoder.
\newblock {\em Applied Sciences}, 11(19):9290, 2021.

\bibitem{cheifetz2011pattern}
Nicolas Cheifetz, Allou Sam{\'e}, Patrice Aknin, and Emmanuel De~Verdalle.
\newblock A pattern recognition approach for anomaly detection on buses brake
  system.
\newblock In {\em 2011 14th International IEEE Conference on Intelligent
  Transportation Systems (ITSC)}, pages 266--271. IEEE, 2011.

\bibitem{bussey2014case}
Howard~E Bussey, Nenad~G Nenadic, Paul~A Ardis, and Michael~G Thurston.
\newblock Case study: Models for detecting low oil pressure anomalies on
  commercial vehicles.
\newblock In {\em Annual Conference of the PHM Society}, volume~6, 2014.

\bibitem{rohrer2018evaluation}
Rodney~A Rohrer, Joe~D Luck, Santosh~K Pitla, and Roger Hoy.
\newblock Evaluation of the accuracy of machine reported can data for engine
  torque and speed.
\newblock {\em Transactions of the ASABE}, 61(5):1547--1557, 2018.

\bibitem{national2011assessment}
National~Research Council et~al.
\newblock {\em Assessment of fuel economy technologies for light-duty
  vehicles}.
\newblock National Academies Press, 2011.

\bibitem{shi2015convolutional}
Xingjian Shi, Zhourong Chen, Hao Wang, Dit-Yan Yeung, Wai-Kin Wong, and
  Wang-chun Woo.
\newblock Convolutional lstm network: A machine learning approach for
  precipitation nowcasting.
\newblock {\em Advances in neural information processing systems}, 28, 2015.

\end{thebibliography}
